\documentclass{article}

\usepackage[preprint]{neurips_2025}

\usepackage[utf8]{inputenc} %
\usepackage[T1]{fontenc}    %
\usepackage{hyperref}       %
\usepackage{url}            %
\usepackage{booktabs}       %
\usepackage{amsfonts}       %
\usepackage{nicefrac}       %
\usepackage{microtype}      %
\usepackage{inconsolata}
\usepackage{caption}
\usepackage{amsmath}
\usepackage{amssymb}
\usepackage{mathtools}
\usepackage{enumitem}
\usepackage{makecell} 
\usepackage[usestackEOL]{stackengine}
\usepackage{graphicx}
\usepackage{capt-of}%
\usepackage{booktabs}
\usepackage{varwidth}
\usepackage[table,dvipsnames,svgnames]{xcolor}

\usepackage{tikz}
\usetikzlibrary{shadows}
\usepackage{float}
\usepackage{caption}
\usepackage{subcaption}
\usepackage{xspace}
\usepackage{svg}

\usepackage{import}
\usepackage{animate}
\usepackage{arydshln}
\usepackage{multirow}
\usepackage[normalem]{ulem}
\usepackage[most]{tcolorbox}
\usepackage{colortbl}
\usepackage{tcolorbox}
\usepackage{pifont}
\usepackage{alltt}
\definecolor{titlegray}{rgb}{0.4, 0.4, 0.4} %
\definecolor{contentgray}{rgb}{0.95, 0.95, 0.95} %
\usepackage{algorithm}
\usepackage{algpseudocode}

\newcommand{\model}{\text{Metis-HOME}\xspace}
\newcommand{\modelbf}{\textbf{Metis-HOME}\xspace}

\definecolor{carolinablue}{rgb}{0.6, 0.73, 0.89}
\definecolor{mildgreen}{rgb}{0.85, 0.98, 0.80}
\definecolor{beautycolor}{rgb}{0.91, 0.75, 0.96} %
\definecolor{fallacycolor}{rgb}{0.85, 0.95, 1}
\definecolor{gendercolor}{rgb}{1, 0.85, 0.85}
\definecolor{brightyellow}{RGB}{255, 255, 100}
\definecolor{boxcolor}{RGB}{51,51,153}
\definecolor{lightgreen}{rgb}{0.56, 0.93, 0.56}
\definecolor{citeblue}{HTML}{0064E0}

\hypersetup{
    colorlinks=true,
    citecolor=citeblue,
    linkcolor=red, %
    urlcolor=citeblue %
}

\definecolor{deepblue}{RGB}{0, 0, 139}

\title{\model: Hybrid Optimized Mixture-of-Experts for Multimodal Reasoning}

\newtcolorbox{questionbanner}{
  colback=blue!10!white,    %
  colframe=blue!80!black,   %
  width=\textwidth,
  arc=4mm,                  %
  boxrule=1pt,              %
  fonttitle=\bfseries,
  title=Question,
}
\newtcolorbox{promptbox}[1]{
  colback=contentgray,      %
  colframe=titlegray,       %
  colbacktitle=titlegray,   %
  coltitle=white,           %
  title={#1}, %
  arc=4mm,                  %
  rounded corners=northwest, %
  rounded corners=northeast, %
  sharp corners=south,      %
  boxrule=1pt,              %
  fonttitle=\bfseries,      %
}

\definecolor{questionbg}{RGB}{240, 248, 255}  %
\definecolor{answerbg}{RGB}{245, 255, 250}   %
\definecolor{bordercolor}{RGB}{100, 149, 237} %
\definecolor{titlecolor}{RGB}{25, 25, 112}    %

\newtcolorbox{vqaexample}[2][]{
    enhanced,
    breakable,
    colback=white,
    colframe=bordercolor,
    boxrule=1.5pt,
    arc=4pt,
    outer arc=4pt,
    left=8pt,
    right=8pt,
    top=8pt,
    bottom=8pt,
    drop shadow={shadow xshift=0.5mm, shadow yshift=-0.5mm, opacity=0.3},
    overlay={
        \node[
            anchor=north east,
            xshift=-3pt,
            yshift=-3pt,
            fill=bordercolor!80,
            text=white,
            font=\bfseries,
            rounded corners=2pt,
            inner sep=4pt,
            minimum height=1.2em,
            align=center
        ] at (frame.north east) {#2};
    },
    #1
}

\author{Xiaohan Lan\thanks{Equal contribution, $^\dagger$Project leader, $^\ddagger$Corresponding author. },\hspace{0.3em}
    Fanfan Liu\footnotemark[1],\hspace{0.3em}
    Haibo Qiu$^\dagger$,\hspace{0.3em}
    Siqi Yang,\hspace{0.3em}
    Delian Ruan,\hspace{0.3em}
    Peng Shi,\hspace{0.3em}
    Lin Ma$^\ddagger$ \\
    \\
    \textbf{Meituan} \\
}

\begin{document}

\maketitle
\vspace{-5mm}
\begin{abstract}
Inspired by recent advancements in LLM reasoning, the field of multimodal reasoning has seen remarkable progress, achieving significant performance gains on intricate tasks such as mathematical problem-solving. Despite this progress, current multimodal large reasoning models exhibit two key limitations. They tend to employ computationally expensive reasoning even for simple queries, leading to inefficiency. Furthermore, this focus on specialized reasoning often impairs their broader, more general understanding capabilities. In this paper, we propose \modelbf: a \textbf{H}ybrid \textbf{O}ptimized \textbf{M}ixture-of-\textbf{E}xperts framework designed to address this trade-off. \model enables a ``Hybrid Thinking'' paradigm by structuring the original dense model into two distinct expert branches: a thinking branch tailored for complex, multi-step reasoning, and a non-thinking branch optimized for rapid, direct inference on tasks like general VQA and OCR. A lightweight, trainable router dynamically allocates queries to the most suitable expert. We instantiate \model by adapting the Qwen2.5-VL-7B into an MoE architecture. Comprehensive evaluations reveal that our approach not only substantially enhances complex reasoning abilities but also improves the model's general capabilities, reversing the degradation trend observed in other reasoning-specialized models. Our work establishes a new paradigm for building powerful and versatile MLLMs, effectively resolving the prevalent reasoning-vs-generalization dilemma. Code and weights are available at \url{https://github.com/MM-Thinking/Metis-HOME}.
\end{abstract}

\begin{figure}[htbp]
    \centering
    \vspace{-5mm}
    \begin{minipage}{0.485\textwidth}
        \centering
        \includegraphics[width=\textwidth]{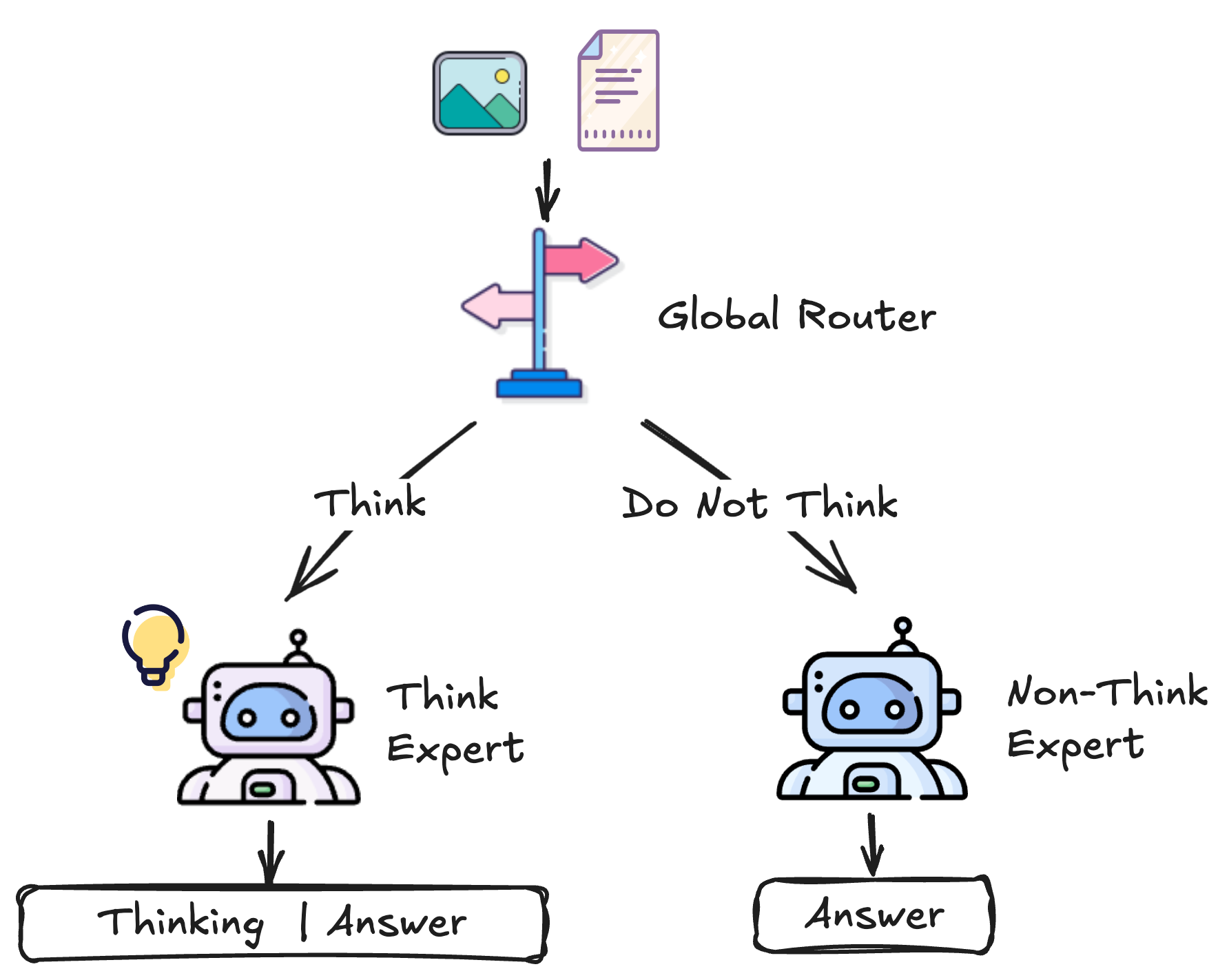}
        \caption{A brief illustration of hybrid thinking paradigm from Meits-HOME.}
        \label{fig:framework}
    \end{minipage}
    \hfill
    \begin{minipage}{0.5\textwidth}
        \centering
        \includegraphics[width=\textwidth]{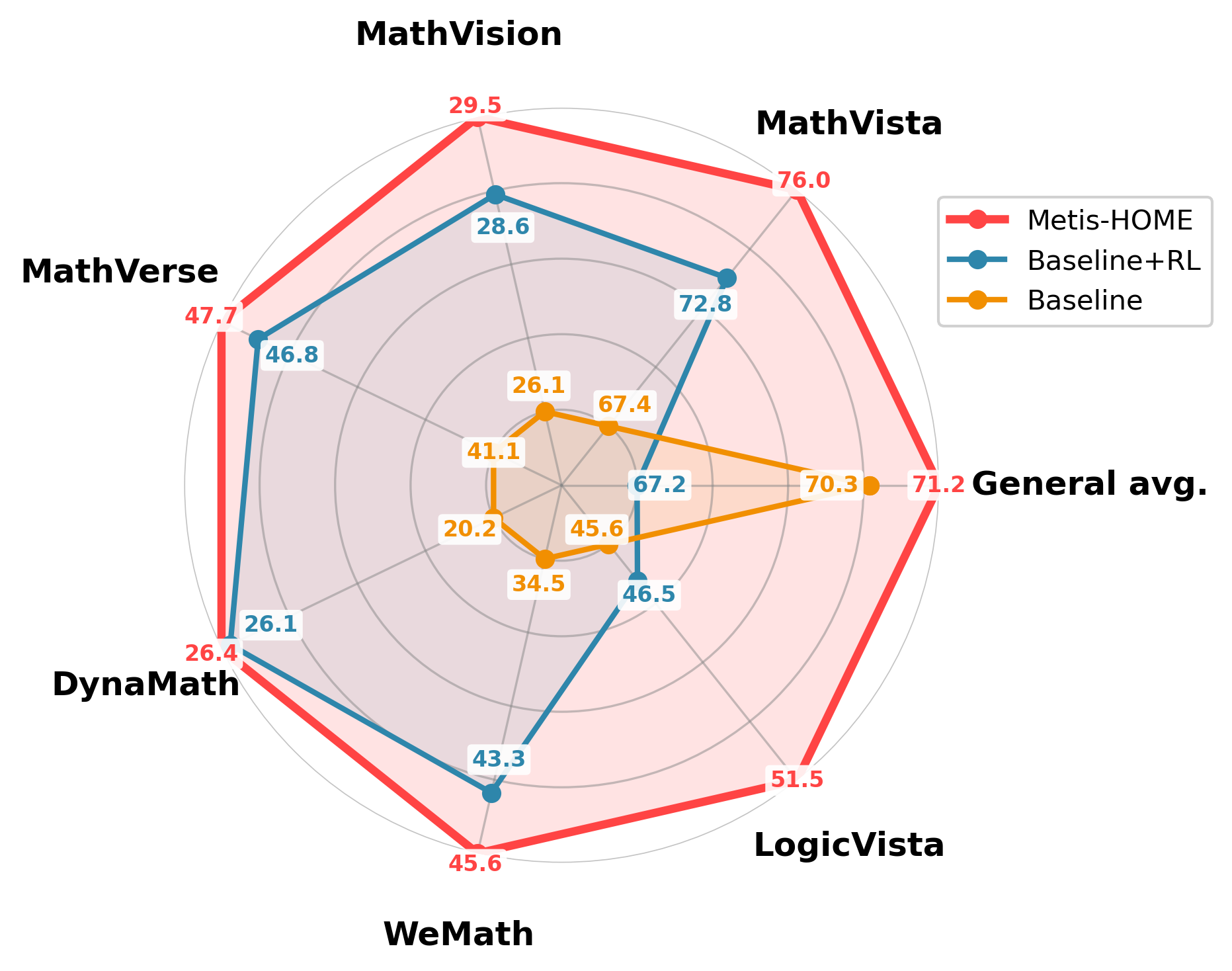}
        \caption{Performance comparisons across different benchmarks of \model against baselines.}
        \label{fig:radar}
    \end{minipage}
\end{figure}
\section{Introduction}
\label{section:introduction}

Recent advances in the complex reasoning of large language models (LLMs) have catalyzed progress in multimodal reasoning. Building on these developments, multimodal reasoning models~\citep{peng2025lmmr1,yang2025r1onevision,shen2025vlm,chen2025r1v,qiu2025metis,meng2025mmeureka,wang2025vl} are achieving notable success on complex tasks such as mathematical problem-solving and scientific question answering.

However, this progress has revealed two key limitations in current multimodal reasoning models. First, they tend to employ computationally expensive reasoning even for simple queries (a phenomenon often termed ``overthinking''), leading to significant inefficiency. Second, the intense focus on specialized mathematical reasoning often impairs the model's broader, general-purpose capabilities. This results in a critical trade-off: enhancing performance on complex reasoning tasks can degrade fundamental skills like general visual question answering (VQA) and optical character recognition (OCR), hindering the development of truly versatile multimodal large language models (MLLMs).

In response to this challenge, recent work has begun to explore the paradigm of hybrid thinking \citep{lou2025adacot,tu2025learning,jiang2025think,zhang2025adaptthink,yang2025qwen3}. The core idea is to enable a model to dynamically switch between a step-by-step thinking mode for complex problems and a direct non-thinking mode for simpler ones. Current implementations range from systems that allow users to manually specify the mode \citep{yang2025qwen3} to models that adaptively determine the appropriate mode based on the query's difficulty~\citep{lou2025adacot,tu2025learning,jiang2025think,zhang2025adaptthink}. For a single, unified model to natively handle both modes, it needs to contend with the significant divergence in output patterns and length distributions between the two modes. Consequently, prior approaches~\citep{jiang2025think,zhang2025adaptthink} have often resorted to designing intricate loss functions or complex reward modeling schemes to balance these competing objectives. While effective to some extent, these methods are often highly sensitive to hyperparameter tuning and add considerable complexity to the training pipeline.

In this paper, we propose \modelbf, as illustrated in Figure~\ref{fig:framework}, a novel \textbf{H}ybrid \textbf{O}ptimized \textbf{M}ixture-of-\textbf{E}xperts framework designed to explicitly resolve the aforementioned dilemma. Our approach explicitly instantiates the hybrid thinking paradigm through a modular MoE architecture. It consists of two specialized expert branches: (1) a \textbf{thinking branch}, fine-tuned for deliberative, multi-step reasoning required for tasks like mathematical problem-solving; and (2) a \textbf{non-thinking branch}, optimized for rapid, direct inference on generalist tasks such as general VQA and OCR. Furthermore, a lightweight and trainable router is positioned at the forefront, dynamically and autonomously dispatching each incoming query to the most suitable expert branch based on its multimodal inputs (i.e., image content and question type) and solving complexity.

We instantiate \model by adapting the widely-used Qwen2.5-VL-7B model~\citep{qwen2_5_vl} into our proposed MoE architecture. The training is conducted via a carefully designed multi-stage strategyz~\citep{qiu2025metis}. First, we employ Reinforcement Learning (RL) to specifically strengthen the model's innate reasoning capabilities, creating the specialized thinking expert. Subsequently, we perform Supervised Fine-Tuning (SFT) using a curated blend of thinking and non-thinking data, and finally obtain the hybrid thinking model. This phased design is motivated by two key insights: (1) complex reasoning abilities require more intensive training to emerge, whereas generalist skills can be effectively recovered or enhanced with a smaller volume of data \citep{dong2023abilities}; (2) the sequence of first enhancing reasoning and then using mixed data to recover generalist capabilities aligns with proven strategies observed in the development of other top-tier models like Qwen3\citep{yang2025qwen3}.

Our comprehensive evaluations highlight the remarkable effectiveness of this strategy. As demonstrated in Figure~\ref{fig:radar}, \model achieves a substantial 6.9\% improvement across six reasoning benchmarks\footnote{\href{https://rank.opencompass.org.cn/leaderboard-multimodal-reasoning/?m=REALTIME}{OpenCompass Multi-modal Reasoning Leaderboard}}, showcasing its enhanced reasoning capabilities. More importantly, it defies the common degradation trend. On the eight comprehensive benchmarks\footnote{\href{https://rank.opencompass.org.cn/leaderboard-multimodal/?m=REALTIME}{OpenCompass Multi-modal Academic Leaderboard}}, \model not only avoids a performance drop but achieves a nearly 1\% gain. This outcome validates our hybrid MoE approach as a successful strategy for resolving the reasoning-vs-generalization dilemma and introduces a new hybrid thinking paradigm for building powerful, versatile MLLMs.

\section{Method}
\label{section:method}
We introduce \model, a novel hybrid optimized mixture-of-experts architecture designed to realize the paradigm of hybrid thinking. Our framework consists of two specialized expert branches, as shown in Figure~\ref{fig:framework}: a \emph{thinking branch}, tailored for deliberative, multi-step reasoning on mathematically intensive and other complex tasks, and a \emph{non-thinking branch}, optimized for rapid and direct inference on conventional tasks such as general VQA and OCR. A lightweight, trainable global router module is deployed at the front-end to automatically assess the type and complexity of each input query, thereby routing it to the most suitable expert branch.

To instantiate this design, we adapt the Qwen2.5-VL-7B model~\citep{qwen2_5_vl} into an MoE architecture. Specifically, for each transformer block in the original dense model, we duplicate and extend the Feed-Forward Network (FFN) into two distinct experts, where one dedicated to thinking and the other to non-thinking processing. A simple yet effective router, constructed with multi-layer perceptrons (MLPs), is integrated to dynamically assign inputs to the appropriate expert. This modular and parameter-efficient expansion preserves the base model’s capacity while enabling adaptive computation based on input complexity, effectively mitigating the trade-off between reasoning depth and generalist performance.

Furthermore, we design a multi-stage training strategy to effectively optimize the proposed framework. First, we employ RL to significantly enhance the original dense model's innate reasoning capabilities. Subsequently, the model is extended into the MoE architecture and undergoes SFT using a blended corpus of carefully curated data from both thinking and non-thinking tasks. This mixed-training phase simultaneously specializes each expert branch and trains the lightweight router, enabling the model to autonomously select between deliberative reasoning and rapid inference based on the content and complexity of each input.

\subsection{Architectural Expansion}
Starting from the original dense model, we expand each transformer block by duplicating the original FFN module into two separate experts: one dedicated to \emph{thinking} and the other to \emph{non-thinking} processing. A \emph{router} module implemented by MLPs is incorporated to dynamically assign inputs to the appropriate expert. 

The use of MoE architectures to distribute processing among specialized modules for different tasks or modalities has also been demonstrated in previous studies. For example, Llava-mole~\citep{chen2024llava} employs multiple experts to implicitly handle diverse task types based on input characteristics. Similarly, Mono-InternVL~\citep{luo2025mono} uses dedicated experts for visual and textual modalities, leveraging modality-specific representations to mitigate inter-modal conflict. In a different setting, WALL-OSS~\citep{zhai2025igniting} assigns separate branches to language and action spaces for more focused modeling. OmniActor~\citep{yang2025omniactor} further illustrates this flexibility with experts designed for 2D GUI and 3D embodied action planning. Together, these works highlight the ability of MoE frameworks to enhance model capacity and specialization while maintaining efficiency.

It is worth noting that we only decouple the FFN modules, while keeping the self-attention and other components shared across experts. This design is motivated by two considerations: (1) it aligns with common practice in MoE architectures~\citep{yang2025qwen3,deepseek-r1}; (2) thinking and non-thinking modes still share substantial common representation learning, particularly in the attention mechanism, which captures contextual relationships essential in both modes. 

We initialize the MoE weights with the RL-trained thinking model based on the insight that reasoning abilities require more intensive and specialized training to emerge, while general capabilities can be efficiently recovered with a smaller amount of data~\citep{dong2023abilities}. This phased approach, first boosting reasoning and later balancing it with generalist skills, is also consistent with strategies adopted in other leading models such as Qwen3 \citep{yang2025qwen3}. Our experimental results in Table~\ref{tab:results_diff_init} of Appendix~\ref{app:general_results} further validate the effectiveness of this initialization strategy.

\subsection{Training Strategy}
Our training strategy unfolds in two main stages. We begin by enhancing the foundational reasoning of the base dense model at Stage-RL. This reasoning-enhanced model is then transformed into our MoE architecture for the second Stage-SFT. During SFT, we use a carefully curated mixed dataset to concurrently train the specialized experts and the router. This process equips our model with the ability to automatically switch between complex reasoning and direct inference based on the input.

\subsubsection{Stage-RL}
Following the strategy of Metis-RISE~\citep{qiu2025metis}, we adapt the Group Relative Policy Optimization (GRPO)~\citep{shao2024deepseekmath} algorithm with advanced optimization techniques from DAPO~\citep{dapo} and VAPO~\citep{yue2025vapo}. Given a query-answer pair $(q, a)$ sampled from the data pool $\mathcal{D}$, the behavior policy model $\pi_{\theta_\text{old}}$ generates a group of $G$ candidate trajectories $\{ \tau_i \}_{i=1}^G$. The objective of RL can be mathematically defined as follows:
\begin{equation}
\begin{aligned}
\mathcal{J}_{\text{RL}}(\theta) =\quad& \mathbb{E}_{(q,a)\sim \mathcal{D}, \{\tau_i\}_{i=1}^G\sim \pi_{\theta_\text{old}}(\cdot\mid q)}\\&
\Bigg[\frac{1}{ \sum_{i=1}^{G}|\tau_i|}{ \sum_{i=1}^{G}\sum_{t=1}^{|\tau_i|}} 
\min \Big( r_{i,t}(\theta) \hat{A}_{i,t},  
\ \text{clip} \Big( r_{i,t}(\theta), 1 - { \varepsilon_{\text{low}}}, 1 + { \varepsilon_{\text{high}}} \Big) \hat{A}_{i,t} \Big) \Bigg]
\\
\text{s.t.}\quad& { 0< \Big|\{\tau_i\mid\texttt{is\_equal}(gt,\tau_i)\}\Big|< G}, 
\label{eq:dapoloss_oversample_filter}
\end{aligned}
\end{equation}
where the importance ratio and advantage are calculated as:
\begin{equation}
    r_{i,t}(\theta)=\frac{\pi_{\theta}(\tau_{i,t} \mid q, \tau_{i,<t})}{\pi_{\theta_{\text{old}}}(\tau_{i,t} \mid q,\tau_{i,<t})},\quad\hat{A}_{i,t} = \frac{R_i - \text{mean}(\{R_i\}_{i=1}^G)}{\text{std}(\{R_i\}_{i=1}^G)}.
\label{eq:advantage_calculation}
\end{equation}

We employ verifiable data during training, enabling automated correctness evaluation via a rule-based verifier. Following DeepSeek-R1~\citep{deepseek-r1}, we adopt a hybrid reward mechanism combining format and accuracy rewards. The format reward enforces strict adherence to a predefined output structure: the model must place its reasoning process within \verb|<think>| and \verb|</think>|, and the final answer within \verb|<answer>| and \verb|</answer>|. Failure to comply results in a zero reward. The accuracy reward is binary (1 or 0) and is assigned only if the extracted answer is verified as correct by the rule-based verifier, thus promoting both structured reasoning and semantic accuracy. The detailed system prompt used for training can be found in Appendix~\ref{app:training}. This phase is aimed at strengthening the model's capacity for complex, multi-step reasoning, providing a solid foundation for the subsequent specialization of the experts.

\subsubsection{Stage-SFT}
\label{section:stage_sft}
In this stage, we perform the mixed SFT using meticulously constructed datasets for both thinking and non-thinking modes to specialize the corresponding experts and train the router. The corresponding system prompt employed for training can be referred in Appendix~\ref{app:training}.

\textbf{Thinking SFT Data}: We follow the data construction protocol of Metis-RISE~\citep{qiu2025metis}. We collect prompts from mathematical and other complex reasoning domains. For each prompt, the model generates $N$ responses (empirically set to 8). The passrate is then defined as the proportion of these responses that are correct. Prompts with a passrate of 1 are discarded, as the model is considered to have fully mastered them. For prompts with passrate between 0 and 1, we use the model’s own correct reasoning trajectories as supervised signals. For those with passrate 0, where the model fails to generate correct responses, we leverage external expert models~\citep{doubao_1_5_pro} to provide reference reasoning trajectories, thereby injecting necessary knowledge into the model. The data format is ``\verb|<think>|\verb|thinking_content|\verb|</think>|\verb|<answer>|\verb|answer_content|\verb|</answer>|''

\textbf{Non-Thinking SFT Data}: We assemble a high-quality dataset comprising general VQA, OCR, and image captioning examples. The volume of this dataset is kept comparable to that of the thinking data to ensure balanced training. In addition, we reformat the data as ``\verb|<think>|\verb|</think>|\verb|<answer>|\verb|answer_content|\verb|</answer>|''

Both types of data are used to learn their respective experts and the router. The training objective combines two cross-entropy losses: one for the final answer prediction and one for the router’s assignment decision. We empirically observe that these two losses are of comparable magnitude and thus simply combine them with a ratio $1:1$ as the total training loss:

\begin{equation}
    \mathcal{L}_{\text{total}} = \mathcal{L}_{\text{prediction}} + \mathcal{L}_{\text{router}}.
    \label{eq:sft_loss}
\end{equation}

This approach enables the model to not only produce accurate responses but also learn to make effective routing decisions in an end-to-end manner.

\section{Experiment}
\label{section:experiment}
In this section, we detail our experimental setup and results. We first describe the implementation of \model and the evaluation protocols. Then we present comprehensive quantitative results. Finally, we provide qualitative case studies to offer intuitive insights into our model's behavior.

\subsection{Implementation and Evaluation}

We build \model upon the open-source Qwen2.5-VL-7B model~\citep{qwen2_5_vl} and extend each transformer block by duplicating the original FFN module into two distinct experts: one dedicated to \emph{thinking} processes and the other to \emph{non-thinking} tasks. Additionally, we incorporate a \emph{router} module, implemented using MLPs, to dynamically allocate inputs to the appropriate expert

In the RL stage, the model is trained with around 40K multimodal reasoning samples by optimizing the objective defined in Equation~\ref{eq:dapoloss_oversample_filter}. For the following SFT stage, we curate around 16K training samples, including 8K thinking samples following Metis-RISE~\citep{qiu2025metis} and 8K non-thinking samples consisting of high-quality general VQA, OCR and caption, and train the model by optimizing the loss defined in Equation~\ref{eq:sft_loss}.

Using VLMEvalKit~\citep{duan2024vlmevalkit}, we comprehensively evaluate the performance of our \model in terms of both reasoning and general capabilities. Six reasoning benchmarks on the \href{https://rank.opencompass.org.cn/leaderboard-multimodal-reasoning/?m=REALTIME}{OpenCompass Multi-modal Reasoning Leaderboard} are adopted for evaluating multimodal mathematical and logical reasoning capabilities, including MathVista~\citep{mathvista}, MathVision~\citep{mathvision}, MathVerse~\citep{mathverse}, DynaMath~\citep{dynamath}, WeMath~\citep{qiao2024we} and LogicVista~\citep{xiao2024logicvista}. For general ability, we employ \href{https://rank.opencompass.org.cn/leaderboard-multimodal/?m=REALTIME}{OpenCompass Multi-modal Academic Leaderboard}, which aggregates eight different and representative multimodal benchmarks, consisting of MMBench~\citep{liu2024mmbench}, MMStar~\citep{chen2024mmstar}, MMMU~\citep{yue2024mmmu}, MathVista~\citep{mathvista}, HallusionBench~\citep{guan2024hallusionbench}, AI2D~\citep{ai2d}, OCRBench~\citep{liu2024ocrbench} and MMVet~\citep{mmvet}.

We compare our \model against two categories of state-of-the-art models: (1) \emph{Proprietary Models}: Gemini-2.0-Pro~\citep{Gemini2}, ChatGPT-4o-20241120~\citep{gpt4o}, Gemini-2.0-Flash~\citep{Gemini2}, and Claude 3.7 Sonnet~\citep{Claude37}; (2) \emph{Open-source Models}: Kimi-VL-A3B-Instruct~\citep{kimiteam2025kimivltechnicalreport}, Qwen2.5-VL-7B~\citep{qwen2_5_vl}, InternVL3-8B~\citep{zhu2025internvl3} and Metis-RISE-7B~\citep{qiu2025metis}.

\subsection{Quantitative Results}
Our quantitative analysis is twofold. First, we benchmark \model against leading SOTA models across a suite of reasoning and general benchmarks. Second, we perform the thinking ratio analysis to evaluate the effectiveness and efficiency of the adaptive and hybrid reasoning mechanism.
\subsubsection{Comparisons Against SOTA}
\begin{table*}[t]
\centering
\caption{Performance comparison of our \model models against prominent proprietary and other open-source models across reasoning and general benchmarks. Baseline indicates Qwen2.5-VL-7B model~\citep{qwen2_5_vl}. Individual scores for each general benchmarks can be found in Table~\ref{tab:general_results} of Appendix~\ref{app:general_results}.}
\resizebox{\textwidth}{!}{
\setlength{\tabcolsep}{3pt}
\begin{tabular}{lcccccccc}
\toprule
\multirow{2}{*}{\textbf{Model}} & \multicolumn{7}{c}{\textbf{Reasoning}}  & \multicolumn{1}{c}{\textbf{General}} \\
\cmidrule(lr){2-8} \cmidrule(lr){9-9}
 & \textbf{MathVista} & \textbf{MathVision} & \textbf{MathVerse} & \textbf{DynaMath} & \textbf{WeMath} & \textbf{LogicVista} & \textbf{Avg.} & \textbf{Avg.} \\
\midrule
\rowcolor{gray!20}
\multicolumn{9}{c}{\textit{\textbf{Proprietary Models}}} \\
Gemini-2.0-Pro & 71.3 & 48.1 & 67.3 & 43.3 & 56.5 & 53.2 & 56.6 & 73.3\\
Gemini-2.0-Flash & 70.4 & 43.6 & 47.8 & 42.1 & 47.4 & 52.3 & 50.6 & 72.6 \\
Claude 3.7 Sonnet & 66.8 & 41.9 & 46.7 & 39.7 & 49.3 & 58.2 & 50.4 & 70.1 \\
ChatGPT-4o & 60.0 & 31.2 & 40.6 & 34.5 & 45.8 & 52.8 & 44.2 & 72.0\\
\midrule
\rowcolor{gray!20}
\multicolumn{9}{c}{\textit{\textbf{Open-source Models}}} \\
LLaVA-OneVision-72B & 67.1 & 25.3 & 27.2 & 15.6 & 32.0& 40.9 & 34.7 & 68.0\\
Kimi-VL-A3B-Instruct & 66.0 & 21.8 & 34.1 & 18.0 & 32.3 & 42.7 & 35.8 & 69.1\\
InternVL3-8B & 70.5 & 30.0 & 38.5 & 25.7 & 39.5 & 44.5 & 41.4 & 73.6\\
VL-Rethinker-7B & 75.5 & 29.3 & 47.2 & 25.4 & 37.8 & 47.0 & 43.7 & 68.3 \\
Metis-RISE-7B & 75.8 & 28.7 & 51.0 & 27.7 & 45.2 & 49.7 & 46.4 & 68.4\\
\midrule
Baseline & 67.4 & 26.2 & 41.1 & 20.2 & 34.5 & 45.6 & \colorbox{yellow!20}{39.2} & 70.3\\
Baseline+RL & 72.8 & 28.7 & 46.8 & 26.2 & 43.3 & 46.5 & 44.0 & \colorbox{blue!10}{67.2}\\
\modelbf & 76.0 & 29.5 & 47.7 & 26.4 & 45.6 & 51.5 & \colorbox{yellow!20}{\textbf{46.1}} & \colorbox{blue!10}{\textbf{71.2}}\\
\bottomrule
\end{tabular}}
\label{tab:main_results}
\end{table*}

Our comprehensive evaluation, as detailed in Table~\ref{tab:main_results}, demonstrates the remarkable effectiveness of \model. On the six reasoning benchmarks, \model achieves an average score of \textbf{46.1\%}, representing a substantial \textbf{6.9\%} absolute improvement over the Baseline (Qwen2.5-VL-7B). This significant gain underscores the success of our specialized thinking branch and training strategy in enhancing complex reasoning capabilities, positioning it competitively among other top-tier reasoning-optimized models. Notably, \model outperforms the dedicated reasoning model VL-Rethinker-7B (43.7\%) and matches the performance of our own previous reasoning specialist, Metis-RISE-7B (46.4\%).

The critical advantage of our hybrid MoE approach, however, is most evident in the preservation of generalist capabilities, an area where specialized models typically suffer. As the results clearly show, both VL-Rethinker-7B and Metis-RISE-7B, despite their strong reasoning performance, experience a substantial degradation in general ability, scoring only \textbf{68.3\%} and \textbf{68.4\%} respectively. This represents a significant drop from the Baseline's \textbf{70.3\%}, confirming the reasoning-vs-generalization trade-off that plagues specialized architectures. Similarly, our Baseline+RL experiment, which applied reinforcement learning to boost reasoning, followed this pattern, incurring a \textbf{3.1\%} drop in general performance (to \textbf{67.2\%}).

However, \model breaks this trend. It not only avoids the degradation seen in other reasoning specialists but actually achieves a \textbf{0.9\%} overall gain, attaining a \textbf{71.2\%} average score on the eight general benchmarks. This result demonstrates that our hybrid framework successfully resolves the dilemma: by dynamically routing queries to a specialized thinking expert or a generalist non-thinking expert, \model delivers superior reasoning prowess without sacrificing, and even slightly enhancing its versatile, generalist abilities. This establishes a new paradigm for developing powerful and balanced multimodal models.

\subsubsection{Thinking Ratio Analysis}
\begin{figure}[htbp]
    \centering
    \includegraphics[width=\linewidth]{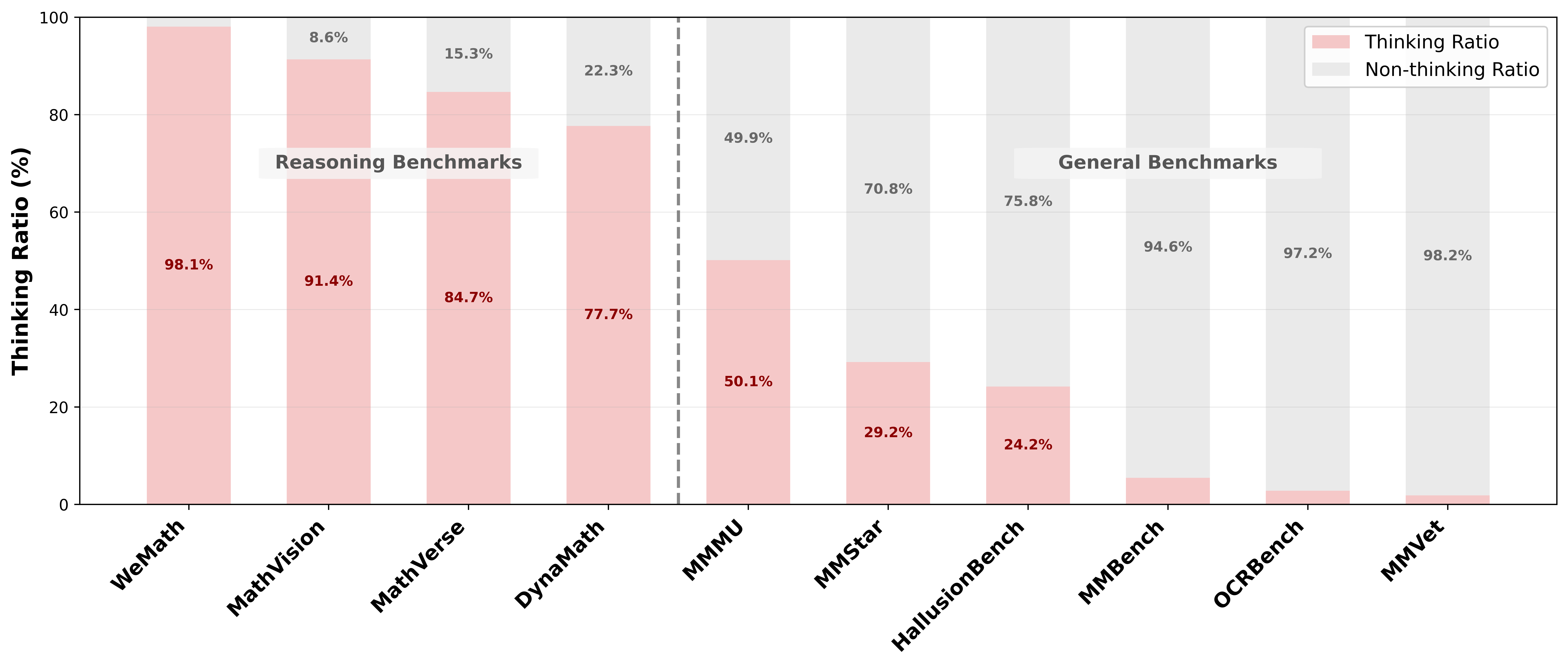}
    \caption{Thinking ratio of \model across reasoning and general benchmarks.}
    \label{fig:thinking_ratio}
\end{figure}

As illustrated in Figure~\ref{fig:thinking_ratio}, we analyze the thinking ratio of \model across a diverse set of benchmarks, which clearly demonstrates the model's adaptive routing behavior. On reasoning-intensive benchmarks such as WeMath, MathVision, MathVerse, and DynaMath, the thinking ratios are notably high (ranging from approximately \textbf{78\%} to \textbf{98\%}), indicating that the router effectively identifies complex queries and directs them to the thinking expert for multi-step reasoning. In contrast, on more general benchmarks like MMBench, OCRBench, and MMVet, the thinking ratios drop significantly (as low as $\sim$\textbf{2\%}–\textbf{5\%}), showing a strong preference for the non-thinking expert. This aligns perfectly with our design intention: the model conserves computational resources by applying deliberative reasoning only when necessary, while resorting to fast and direct inference for simpler tasks. 

Notably, the MMMU benchmark, a comprehensive subject-level evaluation, exhibits an overall thinking ratio of \textbf{50.1\%}, situating it between purely reasoning and general tasks. A fine-grained analysis reveals that this ratio stems from the aggregation of highly divergent subjects: the thinking ratio drops to \textbf{0\%} for topics like ``American Literature'' and ``Fashion Design'', which typically require minimal reasoning, while it reaches \textbf{100\%} for more analytical subjects such as ``Graph Theory'' and ``Calculus''. This further validates our router's ability to identify query complexity not only across broad benchmarks but also within a multidisciplinary dataset, making context-sensitive routing decisions that align with human intuition. These results collectively underscore the effectiveness of our lightweight router and the overall hybrid MoE architecture in achieving intelligent, task-aware computation allocation.

\begin{figure}[htbp]
    \centering
    \begin{minipage}{0.49\textwidth}
        \centering
        \includegraphics[width=\textwidth]{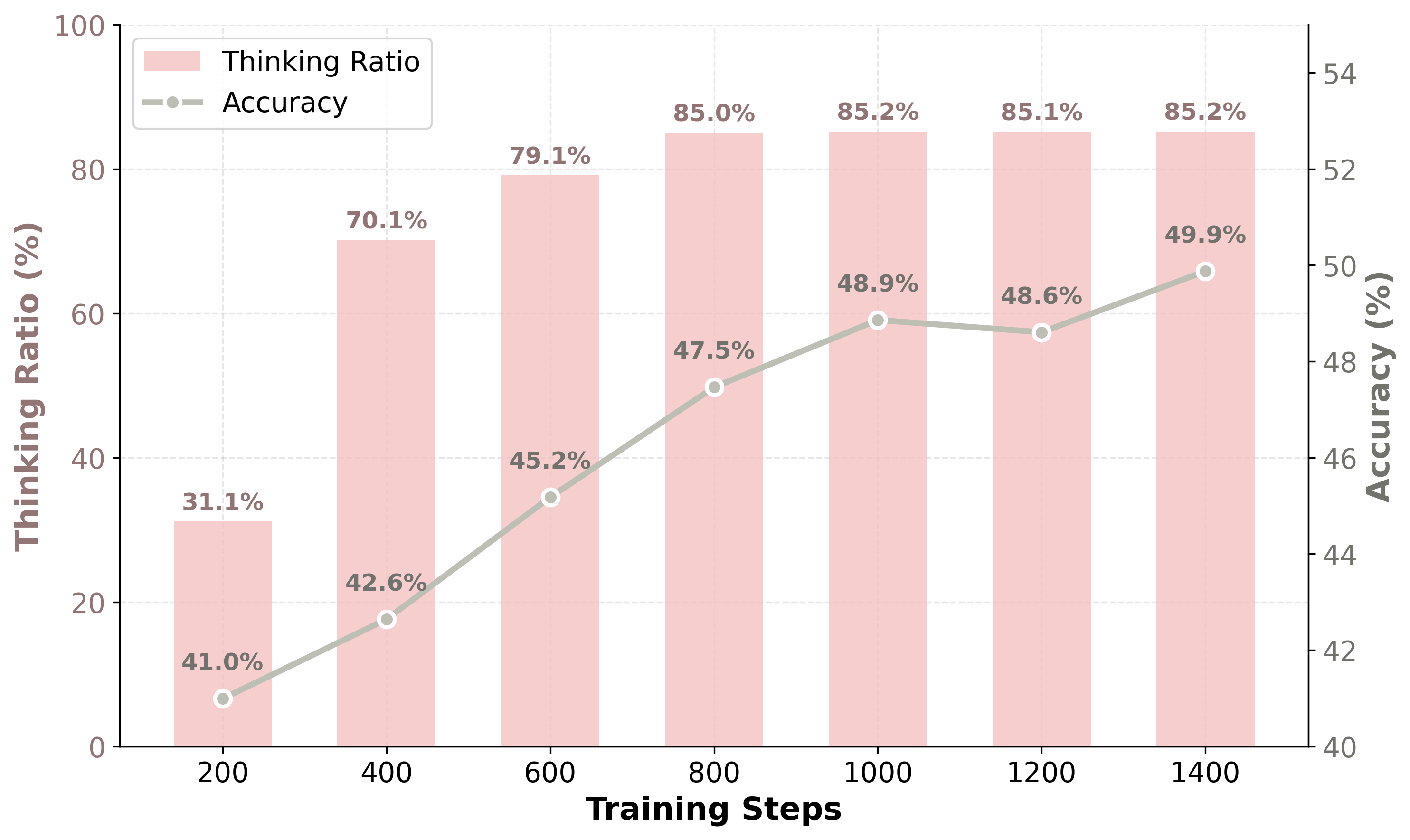}
        \caption{\model~learns to assign more questions to its thinking expert, leading to progressively higher accuracy on MathVerse.}
        \label{fig:performance_trend}
    \end{minipage}
    \hfill
    \begin{minipage}{0.49\textwidth}
        \centering
        \includegraphics[width=\textwidth]{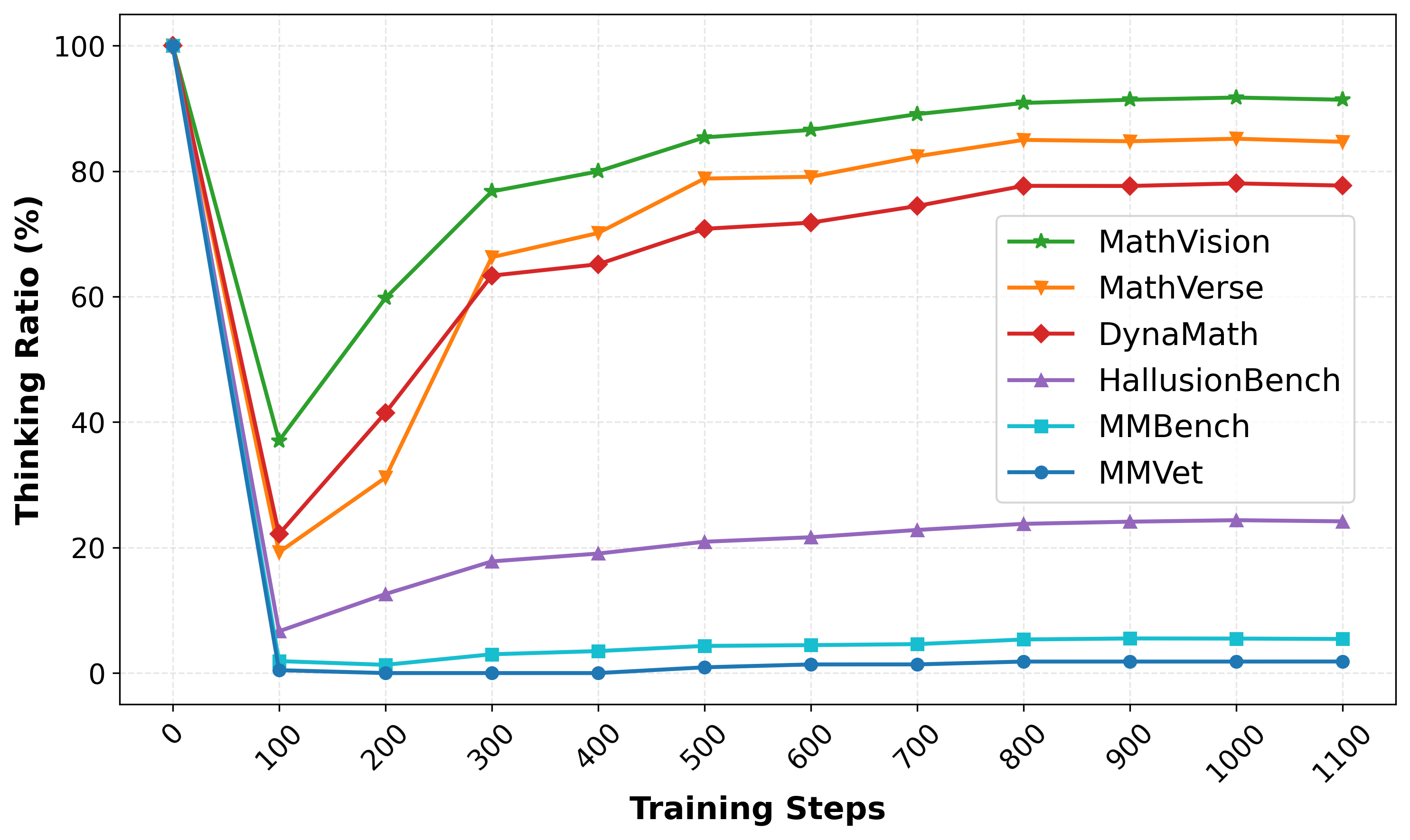}
        \caption{Trend of thinking ratio made by \model during SFT training across different benchmarks.}
        \label{fig:thinking_trend}
    \end{minipage}
\end{figure}

Furthermore, Figure~\ref{fig:performance_trend} illustrates the evolution of both the thinking ratio and corresponding accuracy of \model on the MathVerse benchmark throughout the training process. As the training steps progress, the thinking ratio exhibits a clear upward trend, starting from approximately \textbf{31.1\%} at step 200 and steadily climbing to around \textbf{85.2\%} by step 1400. Notably, this increase in thinking ratio is accompanied by a consistent improvement in accuracy, which rises from \textbf{41.0\%} to nearly \textbf{49.9\%} over the same period. This strong positive correlation demonstrates that \model effectively learns to allocate more queries to the deliberative thinking branch as training advances, and that this increased utilization of multi-step reasoning directly contributes to enhanced performance on complex mathematical tasks. The alignment between the rising thinking ratio and improving accuracy strongly validates our design hypothesis: that encouraging the model to ``think more'' on appropriate problems leads to substantial gains in reasoning capability.

We also illustrate the evolution of the thinking ratio, i.e., the proportion of queries routed to the thinking expert, across various benchmarks throughout the SFT process in Figure~\ref{fig:thinking_trend}. Since the model is initialized from the RL-trained thinking expert, the thinking ratio starts at \textbf{100\%} for all benchmarks at step 0. However, due to the format designs of the thinking and non-thinking data as recalled in Section~\ref{section:stage_sft}, and given that both types are present in a 1:1 ratio, the token \verb|</think>| appears significantly more frequently immediately after the prefix \verb|<think>| across the entire dataset. Under the next-token prediction training paradigm, the model initially prioritizes learning this high-frequency token transition. As a result, early in training, the model tends to quickly overfit to the non-thinking response pattern, leading to a sharp decline in the thinking ratio across all benchmarks.

As training progresses, the model gradually learns to distinguish between tasks that require deliberative reasoning and those that do not. For general-purpose benchmarks such as MMBench and MMVet, the thinking ratio remains consistently low (below \textbf{6\%}), indicating that the model correctly avoids unnecessary reasoning for straightforward visual question-answering tasks. In contrast, for complex mathematical reasoning benchmarks such as MathVision, MathVerse, and DynaMath, the thinking ratio steadily recovers and eventually stabilizes at a high level (above \textbf{77\%}). This demonstrates the router’s improved ability to identify queries that require multi-step reasoning. The router’s adaptive behavior confirms that \model successfully learns a task-aware routing policy: it conserves computational resources on simple queries while robustly activating the thinking expert for challenging problems. This dynamic specialization is key to achieving both high accuracy and efficiency, effectively resolving the reasoning-vs-generalization trade-off.

\subsection{Qualitative Results}

\begin{figure}[!ht]
    \centering
    \begin{vqaexample}[width=\textwidth]{Text Recognition (Non-Thinking)}
        \begin{minipage}[t]{0.3\textwidth}
            \centering
            \includegraphics[width=\textwidth]{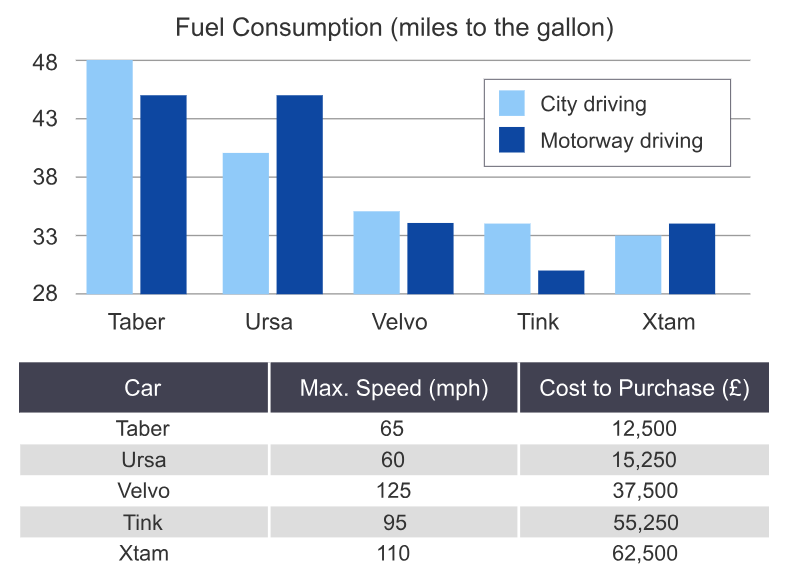}
        \end{minipage}
        \hfill
        \begin{minipage}[t]{0.65\textwidth}
            \begin{tcolorbox}[
                enhanced,
                colback=questionbg,
                colframe=bordercolor!70,
                boxrule=0.8pt,
                arc=3pt,
                left=6pt,
                right=6pt,
                top=4pt,
                bottom=4pt,
                fonttitle=\bfseries\color{titlecolor},
                title=Question: ,
                attach title to upper
            ]
                \ Please recognize the text in the image.
            \end{tcolorbox}
        \end{minipage}

        \begin{tcolorbox}[
            enhanced,
            breakable,
            colback=answerbg,
            colframe=bordercolor!70,
            boxrule=0.8pt,
            arc=3pt,
            left=6pt,
            right=6pt,
            top=4pt,
            bottom=4pt,
            fonttitle=\bfseries\color{titlecolor},
            title=Answer: ,
            attach title to upper
        ]
            \ \colorbox{gray!10}{\texttt{<think>}}
            
            \colorbox{gray!10}{\texttt{</think>}}
            
            \colorbox{gray!10}{\texttt{<answer>}}
            The text in the image is as follows:

            ---

            \textbf{Fuel Consumption (miles to the gallon)}

            \begin{tabular}{|l|c|c|}
            \hline
            Car & Max. Speed (mph) & Cost to Purchase (£) \\
            \hline
            Taber & 65 & 12,500 \\
            Ursa & 60 & 15,250 \\
            Velvo & 125 & 37,500 \\
            Tink & 95 & 55,250 \\
            Xtam & 110 & 62,500 \\
            \hline
            \end{tabular}

            ---

            \textbf{Legend:}

            \begin{itemize}
            \item \textbf{City driving}: Light blue bars
            \item \textbf{Motorway driving}: Dark blue bars
            \end{itemize}
            \colorbox{gray!10}{\texttt{</answer>}}
        \end{tcolorbox}
    \end{vqaexample}
    \caption{Example of a text recognition problem answered by \model.}
    \label{fig:text_recognition_example}
\end{figure}

\begin{figure}[!ht]
    \centering
    \begin{vqaexample}[width=\textwidth]{Image Caption (Non-Thinking)}
        \begin{minipage}[t]{0.15\textwidth}
            \centering
            \includegraphics[width=\textwidth]{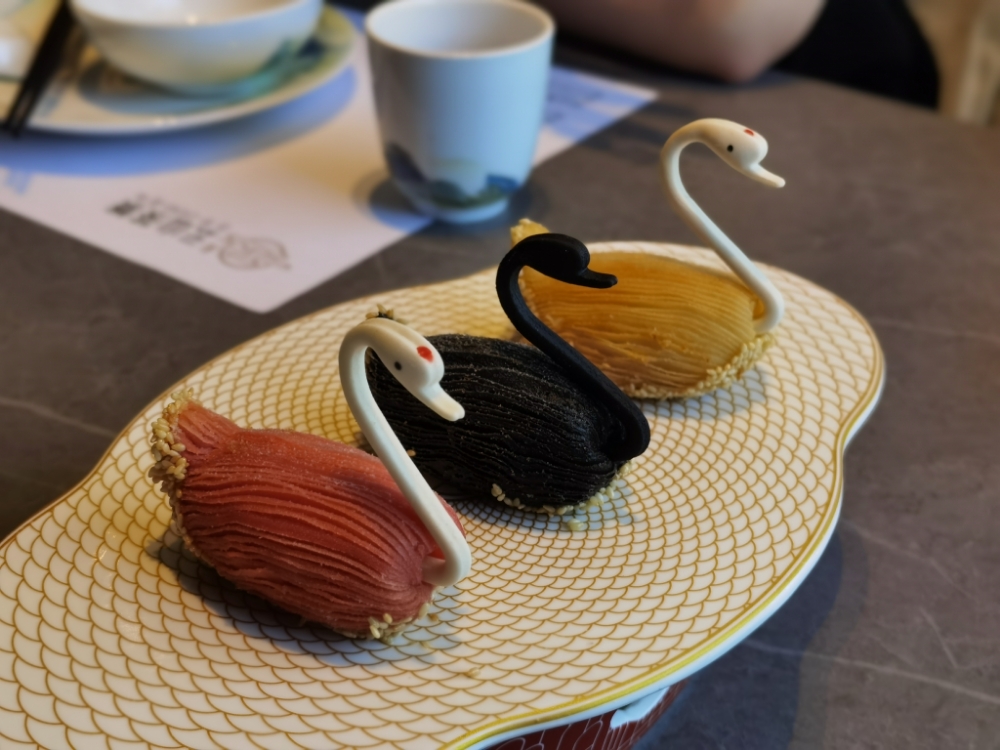}
        \end{minipage}
        \hfill
        \begin{minipage}[t]{0.65\textwidth}
            \begin{tcolorbox}[
                enhanced,
                colback=questionbg,
                colframe=bordercolor!70,
                boxrule=0.8pt,
                arc=3pt,
                left=6pt,
                right=6pt,
                top=4pt,
                bottom=4pt,
                fonttitle=\bfseries\color{titlecolor},
                title=Question: ,
                attach title to upper
            ]
                \ What is it?
            \end{tcolorbox}
        \end{minipage}

        \begin{tcolorbox}[
            enhanced,
            breakable,
            colback=answerbg,
            colframe=bordercolor!70,
            boxrule=0.8pt,
            arc=3pt,
            left=6pt,
            right=6pt,
            top=4pt,
            bottom=4pt,
            fonttitle=\bfseries\color{titlecolor},
            title=Answer: ,
            attach title to upper
        ]
            \ \colorbox{gray!10}{\texttt{<think>}}\colorbox{gray!10}{\texttt{</think>}}

            \colorbox{gray!10}{\texttt{<answer>}}This image shows a plate of decorative swan-shaped pastries. Each swan has a unique color: one is red, another is black, and the third is yellow. These pastries appear to be made from layers of cake or pastry, possibly filled with cream or fruit. They are presented on a plate with a fish-scale pattern, adding an elegant touch to the presentation.\colorbox{gray!10}{\texttt{</answer>}}
        \end{tcolorbox}
    \end{vqaexample}

    \begin{vqaexample}[width=\textwidth]{Plane Geometry (Thinking)}
        \begin{minipage}[t]{0.25\textwidth}
            \centering
            \includegraphics[width=\textwidth]{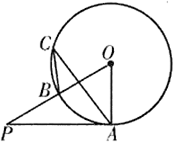}
        \end{minipage}
        \hfill
        \begin{minipage}[t]{0.65\textwidth}
            \begin{tcolorbox}[
                enhanced,
                colback=questionbg,
                colframe=bordercolor!70,
                boxrule=0.8pt,
                arc=3pt,
                left=6pt,
                right=6pt,
                top=4pt,
                bottom=4pt,
                fonttitle=\bfseries\color{titlecolor},
                title=Question: ,
                attach title to upper
            ]
                \ As shown in the figure, $PA$ is tangent to $\odot O$ at point $A$, $PO$ intersects $\odot O$ at point $B$, point $C$ is on the major arc $AB$, and $AC$ and $BC$ are connected. If $\angle P = \angle C$, and the radius of $\odot O$ is 1, then the length of the minor arc $AB$ is. (Express the result in terms of $\pi$)
            \end{tcolorbox}
        \end{minipage}

        \begin{tcolorbox}[
            enhanced,
            breakable,
            colback=answerbg,
            colframe=bordercolor!70,
            boxrule=0.8pt,
            arc=3pt,
            left=6pt,
            right=6pt,
            top=4pt,
            bottom=4pt,
            fonttitle=\bfseries\color{titlecolor},
            title=Answer: ,
            attach title to upper
        ]
            \ \colorbox{gray!10}{\texttt{<think>}}
            To solve the problem, we need to determine the measure of the central angle subtending the arc $AB$. Let's start by noting some properties and relationships in the circle.
            \\

            Since $PA$ is tangent to $\odot O$ at point $A$, $OA$ is perpendicular to $PA$, meaning $\angle OAP = 90^\circ$. Therefore, $\angle P + \angle AOP = 90^\circ$.
            \\

            Given that $\angle P = \angle C$ and $\angle C$ is an inscribed angle that subtends arc $AB$, $\angle AOB$ (the central angle subtending arc $AB$) is twice $\angle C$. Thus, $\angle AOB = 2 \angle C$.
            \\

            Let $\angle C = x$. Then $\angle AOB = 2x$. Since $\angle P = x$ and $\angle P + \angle AOP = 90^\circ$, we have $x + \angle AOP = 90^\circ$. But $\angle AOP = \angle AOB = 2x$, so $x + 2x = 90^\circ$, which gives $3x = 90^\circ$ and thus $x = 30^\circ$. Therefore, $\angle AOB = 60^\circ$.
            \\

            The length of the arc $AB$ can be calculated using the formula for the arc length, $s = r\theta$, where $r$ is the radius and $\theta$ is the central angle in radians. Here, $r = 1$ and $\theta = \frac{\pi}{3}$ (since $60^\circ = \frac{\pi}{3}$ radians).
            \\

            Thus, the length of the arc $AB$ is:
            $$s = 1 \cdot \frac{\pi}{3} = \frac{\pi}{3}$$
            \\

            So, the length of the minor arc $AB$ is $\boxed{\frac{\pi}{3}}$.
            \colorbox{gray!10}{\texttt{</think>}}
            \\
            
            \colorbox{gray!10}{\texttt{<answer>}}
            The answer is $\boxed{\frac{\pi}{3}}$
            \colorbox{gray!10}{\texttt{</answer>}}
        \end{tcolorbox}
    \end{vqaexample}
    \caption{Examples of the image caption and geometry problem solved by \model.}
    \label{fig:caption_geo_example}
\end{figure}

The provided examples vividly demonstrate the core operational principle and efficacy of the \model framework. As illustrated in Figures~\ref{fig:text_recognition_example} and~\ref{fig:caption_geo_example}, the model's lightweight router successfully distinguishes between queries requiring different cognitive loads, dynamically activating the most suitable expert branch.

In the first case (Figure~\ref{fig:text_recognition_example}), an OCR task, which is fundamentally a perception-oriented problem, is seamlessly handled by the \emph{non-thinking branch}. The model accurately extracts and formats the structured textual data from the image without attempting to inject any interpretive reasoning, highlighting the branch's proficiency in tasks where precision and speed are paramount.

The second example (Figure~\ref{fig:caption_geo_example}, top) further illustrates the router’s ability to capture task nature. A straightforward image captioning query (\textit{``What is it?''}) is routed to the \emph{non-thinking branch}. The response is direct, concise, and descriptive, generated without any reasoning steps. This exemplifies the branch's optimization for rapid, accurate inference on generalist tasks, effectively mitigating the \textit{overthinking} problem by avoiding unnecessary computational overhead on simple queries.

Conversely, the third case (Figure~\ref{fig:caption_geo_example}, bottom) presents a complex plane geometry problem. The router correctly identifies this as a task demanding deliberative reasoning and activates the \emph{thinking branch}. The response is characterized by a multi-step, chain-of-thought process. It meticulously decomposes 

\clearpage
\newpage

\newpage

the problem, applies geometric theorems, performs algebraic manipulations, and finally arrives at the precise answer. This showcases the branch's specialized capability for handling intricate multi-step reasoning tasks that are beyond the scope of a direct inference approach.

Collectively, these examples provide concrete evidence that \model successfully operationalizes the hybrid thinking paradigm. The trainable router learns an effective routing policy, ensuring computational resources are allocated efficiently: simple queries bypass expensive reasoning, while complex problems receive the necessary deliberative power. This intrinsic specialization of experts, combined with dynamic routing, is the key mechanism behind \model's ability to enhance performance on complex reasoning benchmarks \textit{while simultaneously} recovering or even improving generalist capabilities, thereby resolving the reasoning-vs-generalization dilemma.

\section{Related Work}
\label{section:related}

\paragraph{Multimodal Reasoning.}
Recent progress in multimodal reasoning has been significantly influenced by advances in reasoning LLMs~\citep{deepseek-r1,openai-o1,seed2025seed,wu2025generalization}, leading to substantial improvements on complex tasks such as mathematical problem-solving and scientific question answering \citep{vl-rethinker,chen2025sftrlearlyinvestigation,qiu2025metis,peng2025lmmr1,chen2025chart,meng2025mmeureka,chen2025r1v,chen2025breaking,zhang2025deepsketcher}. These models typically employ chain-of-thought or similar multi-step reasoning strategies to tackle difficult problems. However, as these systems become more specialized, they often suffer from computational inefficiency on simpler queries and a noticeable decline in general capabilities: a phenomenon we refer to as the reasoning-vs-generalization dilemma.

\paragraph{Hybrid Thinking Paradigm.}
To address these issues, several recent studies have explored the idea of hybrid thinking, which allows a model to dynamically choose between a deliberative “thinking” mode and a direct “non-thinking” mode. Some approaches rely on manual mode selection by the user \citep{yang2025qwen3}, while others attempt to automatically infer the appropriate mode from input content \citep{lou2025adacot, tu2025learning, jiang2025think, zhang2025adaptthink}. These methods often require carefully designed loss functions or reinforcement learning objectives to balance the two modes within a single model, introducing additional training complexity and potential instability. In contrast, our \model proposes to explicitly structures experts into \emph{thinking} and \emph{non-thinking} branches under a unified multimodal MoE framework, enabling more flexible and efficient reasoning without compromising generalist performance.

\section{Conclusion}
\label{section:conclusion}

This paper introduces \model, a Hybrid Optimized Mixture-of-Experts framework designed to resolve the trade-off between complex reasoning and generalist capabilities in multimodal models. By explicitly instantiating a thinking expert and a non-thinking expert under a trainable router, \model achieves a “Hybrid Thinking” paradigm that dynamically adapts to query complexity. Comprehensive quantitative and qualitative analyses demonstrate that our approach not only significantly improves reasoning performance but also achieves gains on generalist benchmarks, reversing the typical capability trade-off. This demonstrates the effectiveness of hybrid MoE architectures in building versatile and efficient multimodal systems. Future work may explore scaling experts and extending the paradigm to more modalities.

\clearpage
\bibliography{neurips_2025}

\begin{thebibliography}{50}
\providecommand{\natexlab}[1]{#1}
\providecommand{\url}[1]{\texttt{#1}}
\expandafter\ifx\csname urlstyle\endcsname\relax
  \providecommand{\doi}[1]{doi: #1}\else
  \providecommand{\doi}{doi: \begingroup \urlstyle{rm}\Url}\fi

\bibitem[Peng et~al.(2025)Peng, Zhang, Zhang, You, Liu, Zhu, Yang, Xu, Geng, and Yang]{peng2025lmmr1}
Yingzhe Peng, Gongrui Zhang, Miaosen Zhang, Zhiyuan You, Jie Liu, Qipeng Zhu, Kai Yang, Xingzhong Xu, Xin Geng, and Xu~Yang.
\newblock Lmm-r1: Empowering 3b lmms with strong reasoning abilities through two-stage rule-based rl.
\newblock \emph{arXiv preprint arXiv:2503.07536}, 2025.

\bibitem[Yang et~al.(2025{\natexlab{a}})Yang, He, Pan, Jiang, Deng, Yang, Lu, Yin, Rao, Zhu, Zhang, and Chen]{yang2025r1onevision}
Yi~Yang, Xiaoxuan He, Hongkun Pan, Xiyan Jiang, Yan Deng, Xingtao Yang, Haoyu Lu, Dacheng Yin, Fengyun Rao, Minfeng Zhu, Bo~Zhang, and Wei Chen.
\newblock R1-onevision: Advancing generalized multimodal reasoning through cross-modal formalization.
\newblock \emph{arXiv preprint arXiv:2503.10615}, 2025{\natexlab{a}}.

\bibitem[Shen et~al.(2025)Shen, Liu, Li, Fang, Ma, Liao, Shen, Zhang, Zhao, Zhang, Xu, and Zhao]{shen2025vlm}
Haozhan Shen, Peng Liu, Jingcheng Li, Chunxin Fang, Yibo Ma, Jiajia Liao, Qiaoli Shen, Zilun Zhang, Kangjia Zhao, Qianqian Zhang, Ruochen Xu, and Tiancheng Zhao.
\newblock Vlm-r1: A stable and generalizable r1-style large vision-language model.
\newblock \emph{arXiv preprint arXiv:2504.07615}, 2025.

\bibitem[Chen et~al.(2025{\natexlab{a}})Chen, Li, Zhao, Song, and Vinci]{chen2025r1v}
Liang Chen, Lei Li, Haozhe Zhao, Yifan Song, and Vinci.
\newblock R1-v: Reinforcing super generalization ability in vision-language models with less than \$3.
\newblock \url{https://github.com/Deep-Agent/R1-V}, 2025{\natexlab{a}}.
\newblock Accessed: 2025-02-02.

\bibitem[Qiu et~al.(2025)Qiu, Lan, Liu, Sun, Ruan, Shi, and Ma]{qiu2025metis}
Haibo Qiu, Xiaohan Lan, Fanfan Liu, Xiaohu Sun, Delian Ruan, Peng Shi, and Lin Ma.
\newblock Metis-rise: Rl incentivizes and sft enhances multimodal reasoning model learning.
\newblock \emph{arXiv preprint arXiv:2506.13056}, 2025.

\bibitem[Meng et~al.(2025)Meng, Du, Liu, Zhou, Lu, Fu, Han, Shi, Wang, He, Zhang, Luo, Qiao, Zhang, and Shao]{meng2025mmeureka}
Fanqing Meng, Lingxiao Du, Zongkai Liu, Zhixiang Zhou, Quanfeng Lu, Daocheng Fu, Tiancheng Han, Botian Shi, Wenhai Wang, Junjun He, Kaipeng Zhang, Ping Luo, Yu~Qiao, Qiaosheng Zhang, and Wenqi Shao.
\newblock Mm-eureka: Exploring the frontiers of multimodal reasoning with rule-based reinforcement learning.
\newblock \emph{arXiv preprint arXiv:2503.07365}, 2025.

\bibitem[Wang et~al.(2025{\natexlab{a}})Wang, Qu, Huang, Chu, Lin, and Chen]{wang2025vl}
Haozhe Wang, Chao Qu, Zuming Huang, Wei Chu, Fangzhen Lin, and Wenhu Chen.
\newblock Vl-rethinker: Incentivizing self-reflection of vision-language models with reinforcement learning.
\newblock \emph{arXiv preprint arXiv:2504.08837}, 2025{\natexlab{a}}.

\bibitem[Lou et~al.(2025)Lou, Sun, Liang, Qu, Shen, Wang, Li, Yang, and Wu]{lou2025adacot}
Chenwei Lou, Zewei Sun, Xinnian Liang, Meng Qu, Wei Shen, Wenqi Wang, Yuntao Li, Qingping Yang, and Shuangzhi Wu.
\newblock Adacot: Pareto-optimal adaptive chain-of-thought triggering via reinforcement learning.
\newblock \emph{arXiv preprint arXiv:2505.11896}, 2025.

\bibitem[Tu et~al.(2025)Tu, Lin, Zhang, Tian, Li, Lan, and Zhao]{tu2025learning}
Songjun Tu, Jiahao Lin, Qichao Zhang, Xiangyu Tian, Linjing Li, Xiangyuan Lan, and Dongbin Zhao.
\newblock Learning when to think: Shaping adaptive reasoning in r1-style models via multi-stage rl.
\newblock \emph{arXiv preprint arXiv:2505.10832}, 2025.

\bibitem[Jiang et~al.(2025)Jiang, Wu, Huang, Dong, Chi, Dong, Zhang, Lv, Cui, and Wei]{jiang2025think}
Lingjie Jiang, Xun Wu, Shaohan Huang, Qingxiu Dong, Zewen Chi, Li~Dong, Xingxing Zhang, Tengchao Lv, Lei Cui, and Furu Wei.
\newblock Think only when you need with large hybrid-reasoning models.
\newblock \emph{arXiv preprint arXiv:2505.14631}, 2025.

\bibitem[Zhang et~al.(2025{\natexlab{a}})Zhang, Lin, Hou, Feng, and Li]{zhang2025adaptthink}
Jiajie Zhang, Nianyi Lin, Lei Hou, Ling Feng, and Juanzi Li.
\newblock Adaptthink: Reasoning models can learn when to think.
\newblock \emph{arXiv preprint arXiv:2505.13417}, 2025{\natexlab{a}}.

\bibitem[Yang et~al.(2025{\natexlab{b}})Yang, Li, Yang, Zhang, Hui, Zheng, Yu, Gao, Huang, Lv, et~al.]{yang2025qwen3}
An~Yang, Anfeng Li, Baosong Yang, Beichen Zhang, Binyuan Hui, Bo~Zheng, Bowen Yu, Chang Gao, Chengen Huang, Chenxu Lv, et~al.
\newblock Qwen3 technical report.
\newblock \emph{arXiv preprint arXiv:2505.09388}, 2025{\natexlab{b}}.

\bibitem[Bai et~al.(2025)Bai, Chen, Liu, Wang, Ge, Song, Dang, Wang, Wang, Tang, et~al.]{qwen2_5_vl}
Shuai Bai, Keqin Chen, Xuejing Liu, Jialin Wang, Wenbin Ge, Sibo Song, Kai Dang, Peng Wang, Shijie Wang, Jun Tang, et~al.
\newblock Qwen2. 5-vl technical report.
\newblock \emph{arXiv preprint arXiv:2502.13923}, 2025.

\bibitem[Dong et~al.(2023)Dong, Yuan, Lu, Li, Xue, Liu, Wang, Yuan, Zhou, and Zhou]{dong2023abilities}
Guanting Dong, Hongyi Yuan, Keming Lu, Chengpeng Li, Mingfeng Xue, Dayiheng Liu, Wei Wang, Zheng Yuan, Chang Zhou, and Jingren Zhou.
\newblock How abilities in large language models are affected by supervised fine-tuning data composition.
\newblock \emph{arXiv preprint arXiv:2310.05492}, 2023.

\bibitem[Chen et~al.(2024{\natexlab{a}})Chen, Jie, and Ma]{chen2024llava}
Shaoxiang Chen, Zequn Jie, and Lin Ma.
\newblock Llava-mole: Sparse mixture of lora experts for mitigating data conflicts in instruction finetuning mllms.
\newblock \emph{arXiv preprint arXiv:2401.16160}, 2024{\natexlab{a}}.

\bibitem[Luo et~al.(2025)Luo, Yang, Dou, Wang, Liu, Dai, Qiao, and Zhu]{luo2025mono}
Gen Luo, Xue Yang, Wenhan Dou, Zhaokai Wang, Jiawen Liu, Jifeng Dai, Yu~Qiao, and Xizhou Zhu.
\newblock Mono-internvl: Pushing the boundaries of monolithic multimodal large language models with endogenous visual pre-training.
\newblock In \emph{Proceedings of the Computer Vision and Pattern Recognition Conference}, pages 24960--24971, 2025.

\bibitem[Zhai et~al.(2025)Zhai, Liu, Fang, Cai, Ma, Yin, Wang, Zhou, Wang, Shi, et~al.]{zhai2025igniting}
Andy Zhai, Brae Liu, Bruno Fang, Chalse Cai, Ellie Ma, Ethan Yin, Hao Wang, Hugo Zhou, James Wang, Lights Shi, et~al.
\newblock Igniting vlms toward the embodied space.
\newblock \emph{arXiv preprint arXiv:2509.11766}, 2025.

\bibitem[Yang et~al.(2025{\natexlab{c}})Yang, Zeng, Zhong, Huang, Zheng, Chen, Qiu, Qin, Ma, and Li]{yang2025omniactor}
Longrong Yang, Zhixiong Zeng, Yufeng Zhong, Jing Huang, Liming Zheng, Lei Chen, Haibo Qiu, Zequn Qin, Lin Ma, and Xi~Li.
\newblock Omniactor: A generalist gui and embodied agent for 2d\&3d worlds.
\newblock \emph{arXiv preprint arXiv:2509.02322}, 2025{\natexlab{c}}.

\bibitem[Guo et~al.(2025)Guo, Yang, Zhang, Song, Zhang, Xu, Zhu, Ma, Wang, Bi, et~al.]{deepseek-r1}
Daya Guo, Dejian Yang, Haowei Zhang, Junxiao Song, Ruoyu Zhang, Runxin Xu, Qihao Zhu, Shirong Ma, Peiyi Wang, Xiao Bi, et~al.
\newblock Deepseek-r1: Incentivizing reasoning capability in llms via reinforcement learning.
\newblock \emph{arXiv preprint arXiv:2501.12948}, 2025.

\bibitem[Shao et~al.(2024)Shao, Wang, Zhu, Xu, Song, Bi, Zhang, Zhang, Li, Wu, et~al.]{shao2024deepseekmath}
Zhihong Shao, Peiyi Wang, Qihao Zhu, Runxin Xu, Junxiao Song, Xiao Bi, Haowei Zhang, Mingchuan Zhang, YK~Li, Y~Wu, et~al.
\newblock Deepseekmath: Pushing the limits of mathematical reasoning in open language models.
\newblock \emph{arXiv preprint arXiv:2402.03300}, 2024.

\bibitem[Yu et~al.(2025)Yu, Zhang, Zhu, Yuan, Zuo, Yue, Fan, Liu, Liu, Liu, et~al.]{dapo}
Qiying Yu, Zheng Zhang, Ruofei Zhu, Yufeng Yuan, Xiaochen Zuo, Yu~Yue, Tiantian Fan, Gaohong Liu, Lingjun Liu, Xin Liu, et~al.
\newblock Dapo: An open-source llm reinforcement learning system at scale.
\newblock \emph{arXiv preprint arXiv:2503.14476}, 2025.

\bibitem[Yue et~al.(2025)Yue, Yuan, Yu, Zuo, Zhu, Xu, Chen, Wang, Fan, Du, et~al.]{yue2025vapo}
Yu~Yue, Yufeng Yuan, Qiying Yu, Xiaochen Zuo, Ruofei Zhu, Wenyuan Xu, Jiaze Chen, Chengyi Wang, TianTian Fan, Zhengyin Du, et~al.
\newblock Vapo: Efficient and reliable reinforcement learning for advanced reasoning tasks.
\newblock \emph{arXiv preprint arXiv:2504.05118}, 2025.

\bibitem[Seed(2025)]{doubao_1_5_pro}
ByteDance Seed.
\newblock Doubao-1.5-pro, 2025.
\newblock URL \url{https://seed.bytedance.com/en/special/doubao_1_5_pro}.

\bibitem[Duan et~al.(2024)Duan, Yang, Qiao, Fang, Chen, Liu, Dong, Zang, Zhang, Wang, et~al.]{duan2024vlmevalkit}
Haodong Duan, Junming Yang, Yuxuan Qiao, Xinyu Fang, Lin Chen, Yuan Liu, Xiaoyi Dong, Yuhang Zang, Pan Zhang, Jiaqi Wang, et~al.
\newblock Vlmevalkit: An open-source toolkit for evaluating large multi-modality models.
\newblock In \emph{Proceedings of the 32nd ACM International Conference on Multimedia}, pages 11198--11201, 2024.

\bibitem[Lu et~al.(2023)Lu, Bansal, Xia, Liu, Li, Hajishirzi, Cheng, Chang, Galley, and Gao]{mathvista}
Pan Lu, Hritik Bansal, Tony Xia, Jiacheng Liu, Chunyuan Li, Hannaneh Hajishirzi, Hao Cheng, Kai-Wei Chang, Michel Galley, and Jianfeng Gao.
\newblock Mathvista: Evaluating mathematical reasoning of foundation models in visual contexts.
\newblock \emph{arXiv preprint arXiv:2310.02255}, 2023.

\bibitem[Wang et~al.(2024)Wang, Pan, Shi, Lu, Ren, Zhou, Zhan, and Li]{mathvision}
Ke~Wang, Junting Pan, Weikang Shi, Zimu Lu, Houxing Ren, Aojun Zhou, Mingjie Zhan, and Hongsheng Li.
\newblock Measuring multimodal mathematical reasoning with math-vision dataset.
\newblock \emph{Advances in Neural Information Processing Systems}, 37:\penalty0 95095--95169, 2024.

\bibitem[Zhang et~al.(2024)Zhang, Jiang, Zhang, Lin, Guo, Qiu, Zhou, Lu, Chang, Qiao, et~al.]{mathverse}
Renrui Zhang, Dongzhi Jiang, Yichi Zhang, Haokun Lin, Ziyu Guo, Pengshuo Qiu, Aojun Zhou, Pan Lu, Kai-Wei Chang, Yu~Qiao, et~al.
\newblock Mathverse: Does your multi-modal llm truly see the diagrams in visual math problems?
\newblock In \emph{European Conference on Computer Vision}, pages 169--186, 2024.

\bibitem[Zou et~al.(2024)Zou, Guo, Yang, Zhang, Hu, and Zhang]{dynamath}
Chengke Zou, Xingang Guo, Rui Yang, Junyu Zhang, Bin Hu, and Huan Zhang.
\newblock Dynamath: A dynamic visual benchmark for evaluating mathematical reasoning robustness of vision language models.
\newblock In \emph{The Thirteenth International Conference on Learning Representations}, 2024.

\bibitem[Qiao et~al.(2024)Qiao, Tan, Dong, Wu, Sun, Song, GongQue, Lei, Wei, Zhang, et~al.]{qiao2024we}
Runqi Qiao, Qiuna Tan, Guanting Dong, Minhui Wu, Chong Sun, Xiaoshuai Song, Zhuoma GongQue, Shanglin Lei, Zhe Wei, Miaoxuan Zhang, et~al.
\newblock We-math: Does your large multimodal model achieve human-like mathematical reasoning?
\newblock \emph{arXiv preprint arXiv:2407.01284}, 2024.

\bibitem[Xiao et~al.(2024)Xiao, Sun, Liu, and Wang]{xiao2024logicvista}
Yijia Xiao, Edward Sun, Tianyu Liu, and Wei Wang.
\newblock Logicvista: Multimodal llm logical reasoning benchmark in visual contexts.
\newblock \emph{arXiv preprint arXiv:2407.04973}, 2024.

\bibitem[Liu et~al.(2024{\natexlab{a}})Liu, Duan, Zhang, Li, Zhang, Zhao, Yuan, Wang, He, Liu, et~al.]{liu2024mmbench}
Yuan Liu, Haodong Duan, Yuanhan Zhang, Bo~Li, Songyang Zhang, Wangbo Zhao, Yike Yuan, Jiaqi Wang, Conghui He, Ziwei Liu, et~al.
\newblock Mmbench: Is your multi-modal model an all-around player?
\newblock In \emph{European conference on computer vision}, pages 216--233. Springer, 2024{\natexlab{a}}.

\bibitem[Chen et~al.(2024{\natexlab{b}})Chen, Li, Dong, Zhang, Zang, Chen, Duan, Wang, Qiao, Lin, et~al.]{chen2024mmstar}
Lin Chen, Jinsong Li, Xiaoyi Dong, Pan Zhang, Yuhang Zang, Zehui Chen, Haodong Duan, Jiaqi Wang, Yu~Qiao, Dahua Lin, et~al.
\newblock Are we on the right way for evaluating large vision-language models?
\newblock \emph{Advances in Neural Information Processing Systems}, 37:\penalty0 27056--27087, 2024{\natexlab{b}}.

\bibitem[Yue et~al.(2024)Yue, Ni, Zhang, Zheng, Liu, Zhang, Stevens, Jiang, Ren, Sun, et~al.]{yue2024mmmu}
Xiang Yue, Yuansheng Ni, Kai Zhang, Tianyu Zheng, Ruoqi Liu, Ge~Zhang, Samuel Stevens, Dongfu Jiang, Weiming Ren, Yuxuan Sun, et~al.
\newblock Mmmu: A massive multi-discipline multimodal understanding and reasoning benchmark for expert agi.
\newblock In \emph{Proceedings of the IEEE/CVF Conference on Computer Vision and Pattern Recognition}, pages 9556--9567, 2024.

\bibitem[Guan et~al.(2024)Guan, Liu, Wu, Xian, Li, Liu, Wang, Chen, Huang, Yacoob, et~al.]{guan2024hallusionbench}
Tianrui Guan, Fuxiao Liu, Xiyang Wu, Ruiqi Xian, Zongxia Li, Xiaoyu Liu, Xijun Wang, Lichang Chen, Furong Huang, Yaser Yacoob, et~al.
\newblock Hallusionbench: an advanced diagnostic suite for entangled language hallucination and visual illusion in large vision-language models.
\newblock In \emph{Proceedings of the IEEE/CVF Conference on Computer Vision and Pattern Recognition}, pages 14375--14385, 2024.

\bibitem[Kembhavi et~al.(2016)Kembhavi, Salvato, Kolve, Seo, Hajishirzi, and Farhadi]{ai2d}
Aniruddha Kembhavi, Mike Salvato, Eric Kolve, Minjoon Seo, Hannaneh Hajishirzi, and Ali Farhadi.
\newblock A diagram is worth a dozen images, 2016.
\newblock URL \url{https://arxiv.org/abs/1603.07396}.

\bibitem[Liu et~al.(2024{\natexlab{b}})Liu, Li, Huang, Yang, Yu, Li, Yin, Liu, Jin, and Bai]{liu2024ocrbench}
Yuliang Liu, Zhang Li, Mingxin Huang, Biao Yang, Wenwen Yu, Chunyuan Li, Xu-Cheng Yin, Cheng-Lin Liu, Lianwen Jin, and Xiang Bai.
\newblock Ocrbench: on the hidden mystery of ocr in large multimodal models.
\newblock \emph{Science China Information Sciences}, 67\penalty0 (12):\penalty0 220102, 2024{\natexlab{b}}.

\bibitem[Yu et~al.(2024)Yu, Yang, Li, Wang, Lin, Liu, Wang, and Wang]{mmvet}
Weihao Yu, Zhengyuan Yang, Linjie Li, Jianfeng Wang, Kevin Lin, Zicheng Liu, Xinchao Wang, and Lijuan Wang.
\newblock Mm-vet: Evaluating large multimodal models for integrated capabilities.
\newblock In \emph{International Conference on Machine Learning}, pages 57730--57754. PMLR, 2024.

\bibitem[Google(2024)]{Gemini2}
Google.
\newblock Introducing gemini 2.0: our new ai model for the agentic era, 2024.
\newblock URL \url{https://blog.google/technology/google-deepmind/google-gemini-ai-update-december-2024/#ceo-message}.

\bibitem[Hurst et~al.(2024)Hurst, Lerer, Goucher, Perelman, Ramesh, Clark, Ostrow, Welihinda, Hayes, Radford, et~al.]{gpt4o}
Aaron Hurst, Adam Lerer, Adam~P Goucher, Adam Perelman, Aditya Ramesh, Aidan Clark, AJ~Ostrow, Akila Welihinda, Alan Hayes, Alec Radford, et~al.
\newblock Gpt-4o system card.
\newblock \emph{arXiv preprint arXiv:2410.21276}, 2024.

\bibitem[Anthropic(2025)]{Claude37}
Anthropic.
\newblock Claude 3.7 sonnet and claude code, 2025.
\newblock URL \url{https://www.anthropic.com/news/claude-3-7-sonnet}.

\bibitem[Team et~al.(2025)Team, Du, Yin, Xing, Qu, Wang, Chen, Zhang, Du, Wei, Wang, Zhang, Du, Wang, Yuan, Lu, Li, Sung, Wei, Lai, Zhu, Ding, Hu, Yang, Zhang, Wu, Yao, Lu, Wang, Gao, Zheng, Li, Su, Wang, Deng, Qiu, Xie, Wang, Liu, Yan, Ouyang, Chen, Sui, Yu, Dong, Dong, Xu, Cheng, Gu, Zhou, Liu, Cao, Yu, Song, Bai, Song, He, Huang, Xu, Yuan, Yao, Wu, Zu, Zhou, Wang, Charles, Zhong, Li, Hu, Chen, Wang, Liu, Miao, Qin, Chen, Bao, Wang, Kang, Liu, Du, Wu, Wang, Yan, Zhou, Li, Jiang, Zhang, Yang, Huang, Huang, Zhao, and Chen]{kimiteam2025kimivltechnicalreport}
Kimi Team, Angang Du, Bohong Yin, Bowei Xing, Bowen Qu, Bowen Wang, Cheng Chen, Chenlin Zhang, Chenzhuang Du, Chu Wei, Congcong Wang, Dehao Zhang, Dikang Du, Dongliang Wang, Enming Yuan, Enzhe Lu, Fang Li, Flood Sung, Guangda Wei, Guokun Lai, Han Zhu, Hao Ding, Hao Hu, Hao Yang, Hao Zhang, Haoning Wu, Haotian Yao, Haoyu Lu, Heng Wang, Hongcheng Gao, Huabin Zheng, Jiaming Li, Jianlin Su, Jianzhou Wang, Jiaqi Deng, Jiezhong Qiu, Jin Xie, Jinhong Wang, Jingyuan Liu, Junjie Yan, Kun Ouyang, Liang Chen, Lin Sui, Longhui Yu, Mengfan Dong, Mengnan Dong, Nuo Xu, Pengyu Cheng, Qizheng Gu, Runjie Zhou, Shaowei Liu, Sihan Cao, Tao Yu, Tianhui Song, Tongtong Bai, Wei Song, Weiran He, Weixiao Huang, Weixin Xu, Xiaokun Yuan, Xingcheng Yao, Xingzhe Wu, Xinxing Zu, Xinyu Zhou, Xinyuan Wang, Y.~Charles, Yan Zhong, Yang Li, Yangyang Hu, Yanru Chen, Yejie Wang, Yibo Liu, Yibo Miao, Yidao Qin, Yimin Chen, Yiping Bao, Yiqin Wang, Yongsheng Kang, Yuanxin Liu, Yulun Du, Yuxin Wu, Yuzhi Wang, Yuzi Yan, Zaida Zhou, Zhaowei Li, Zhejun
  Jiang, Zheng Zhang, Zhilin Yang, Zhiqi Huang, Zihao Huang, Zijia Zhao, and Ziwei Chen.
\newblock {Kimi-VL} technical report, 2025.
\newblock URL \url{https://arxiv.org/abs/2504.07491}.

\bibitem[Zhu et~al.(2025)Zhu, Wang, Chen, Liu, Ye, Gu, Duan, Tian, Su, Shao, et~al.]{zhu2025internvl3}
Jinguo Zhu, Weiyun Wang, Zhe Chen, Zhaoyang Liu, Shenglong Ye, Lixin Gu, Yuchen Duan, Hao Tian, Weijie Su, Jie Shao, et~al.
\newblock Internvl3: Exploring advanced training and test-time recipes for open-source multimodal models.
\newblock \emph{arXiv preprint arXiv:2504.10479}, 2025.

\bibitem[Jaech et~al.(2024)Jaech, Kalai, Lerer, Richardson, El-Kishky, Low, Helyar, Madry, Beutel, Carney, et~al.]{openai-o1}
Aaron Jaech, Adam Kalai, Adam Lerer, Adam Richardson, Ahmed El-Kishky, Aiden Low, Alec Helyar, Aleksander Madry, Alex Beutel, Alex Carney, et~al.
\newblock Openai o1 system card.
\newblock \emph{arXiv preprint arXiv:2412.16720}, 2024.

\bibitem[Seed et~al.(2025)Seed, Yuan, Yue, Wang, Zuo, Chen, Yan, Xu, Zhang, Liu, et~al.]{seed2025seed}
ByteDance Seed, Yufeng Yuan, Yu~Yue, Mingxuan Wang, Xiaochen Zuo, Jiaze Chen, Lin Yan, Wenyuan Xu, Chi Zhang, Xin Liu, et~al.
\newblock Seed-thinking-v1. 5: Advancing superb reasoning models with reinforcement learning.
\newblock \emph{arXiv preprint arXiv:2504.13914}, 2025.

\bibitem[Wu et~al.(2025)Wu, Zhou, Ziheng, Peng, Ye, Hu, Zhu, Qi, Yang, and Yang]{wu2025generalization}
Yongliang Wu, Yizhou Zhou, Zhou Ziheng, Yingzhe Peng, Xinyu Ye, Xinting Hu, Wenbo Zhu, Lu~Qi, Ming-Hsuan Yang, and Xu~Yang.
\newblock On the generalization of sft: A reinforcement learning perspective with reward rectification.
\newblock \emph{arXiv preprint arXiv:2508.05629}, 2025.

\bibitem[Wang et~al.(2025{\natexlab{b}})Wang, Qu, Huang, Chu, Lin, and Chen]{vl-rethinker}
Haozhe Wang, Chao Qu, Zuming Huang, Wei Chu, Fangzhen Lin, and Wenhu Chen.
\newblock Vl-rethinker: Incentivizing self-reflection of vision-language models with reinforcement learning.
\newblock \emph{arXiv preprint arXiv:2504.08837}, 2025{\natexlab{b}}.

\bibitem[Chen et~al.(2025{\natexlab{b}})Chen, Tu, Wang, Liu, Tang, Du, Zhou, and Xie]{chen2025sftrlearlyinvestigation}
Hardy Chen, Haoqin Tu, Fali Wang, Hui Liu, Xianfeng Tang, Xinya Du, Yuyin Zhou, and Cihang Xie.
\newblock Sft or rl? an early investigation into training r1-like reasoning large vision-language models, 2025{\natexlab{b}}.
\newblock URL \url{https://arxiv.org/abs/2504.11468}.

\bibitem[Chen et~al.(2025{\natexlab{c}})Chen, Zhao, Zeng, Huang, Zhong, and Ma]{chen2025chart}
Lei Chen, Xuanle Zhao, Zhixiong Zeng, Jing Huang, Yufeng Zhong, and Lin Ma.
\newblock Chart-r1: Chain-of-thought supervision and reinforcement for advanced chart reasoner.
\newblock \emph{arXiv preprint arXiv:2507.15509}, 2025{\natexlab{c}}.

\bibitem[Chen et~al.(2025{\natexlab{d}})Chen, Zhao, Zeng, Huang, Zheng, Zhong, and Ma]{chen2025breaking}
Lei Chen, Xuanle Zhao, Zhixiong Zeng, Jing Huang, Liming Zheng, Yufeng Zhong, and Lin Ma.
\newblock Breaking the sft plateau: Multimodal structured reinforcement learning for chart-to-code generation.
\newblock \emph{arXiv preprint arXiv:2508.13587}, 2025{\natexlab{d}}.

\bibitem[Zhang et~al.(2025{\natexlab{b}})Zhang, Qiu, Zhang, Zeng, Ma, and Zhang]{zhang2025deepsketcher}
Chi Zhang, Haibo Qiu, Qiming Zhang, Zhixiong Zeng, Lin Ma, and Jing Zhang.
\newblock Deepsketcher: Internalizing visual manipulation for multimodal reasoning.
\newblock \emph{arXiv preprint arXiv:2509.25866}, 2025{\natexlab{b}}.

\end{thebibliography}
\bibliographystyle{unsrtnat}

\appendix

\appendix
\section{Training Details}
\label{app:training}

The following system prompts are employed during training.
\begin{promptbox}{System Prompt For Thinking During Stage-RL}
\begin{verbatim}
Solve the question. The user asks a question, and you solves it. You 
first thinks about the reasoning process in the mind and then provides 
the user with the answer. The answer is in latex format and wrapped in 
$...$. The final answer must be wrapped using the \\boxed{} command. 
The reasoning process and answer are enclosed within <think> </think> 
and <answer> </answer> tags, respectively, i.e., <think> Since $1+1=2$, 
so the answer is $2$. </think><answer> The answer is $\\boxed{2}$ 
</answer>, which means assistant's output should start with 
<think> and end with </answer>.
\end{verbatim}
\end{promptbox}

\begin{promptbox}{System Prompt For Hybrid Thinking During Stage-SFT}
\begin{verbatim}
You are a helpful assistant that solves questions and provides answers. When 
the user asks a question, you should analyze it and respond appropriately 
based on the question type.
For mathematical or logical reasoning questions with definitive answers:
- First think through the reasoning process within <think> </think> tags
- Then provide the final answer within <answer> </answer> tags
- The definitive answer must be formatted in LaTeX and wrapped using the
\\boxed{} command
For general visual question answering (VQA) or open-ended questions:
- Use empty <think> </think> tags
- Provide your response within <answer> </answer> tags
- No \\boxed{} formatting is required for open-ended answers
Your response should always start with <think> and end with </answer>.
\end{verbatim}
\end{promptbox}

\clearpage
\section{Detailed Benchmarks Results}
\label{app:general_results}

\begin{table}[htbp]
\centering
\caption{Performance comparison of our \model models against prominent proprietary and other open-source models across general benchmarks.}
\label{tab:general_results}
\resizebox{\textwidth}{!}{
\setlength{\tabcolsep}{3pt}
\begin{tabular}{lccccccccc}
\toprule
\textbf{Model} & \textbf{Avg.} & \textbf{MMBench} & \textbf{MMStar} & \textbf{MMMU} & \textbf{MathVista} & \textbf{Hallusion} & \textbf{AI2D} & \textbf{OCRBench} & \textbf{MMVet} \\
\midrule
\rowcolor{gray!20}
\multicolumn{10}{c}{\textit{\textbf{Proprietary Models}}} \\
Gemini-2.0-Pro & 73.3 & 83.0 & 68.5 & 72.6 & 71.3 & 49.8 & 84.8 & 86.3 & 70.4 \\
Gemini-2.0-Flash & 72.6 & 71.0 & 69.4 & 69.9 & 70.4 & 58.0 & 83.1 & 85.1 & 73.6 \\
Claude 3.7 Sonnet & 70.1 & 79.7 & 65.1 & 71.0 & 66.8 & 55.4 & 82.5 & 70.1 & 70.0 \\
ChatGPT-4o & 72.0 & 84.3 & 65.1 & 70.7 & 60.0 & 56.2 & 84.9 & 80.6 & 74.5 \\

\midrule
\rowcolor{gray!20}
\multicolumn{10}{c}{\textit{\textbf{Open-source Models}}} \\
LLaVA-OneVision-72B & 68.0 & 84.5 & 65.8 & 56.6 & 68.4 & 47.9 & 86.2 & 74.1 & 60.6 \\
Kimi-VL-A3B-Instruct & 69.1 & 80.8 & 62.0 & 57.8 & 66.0 & 48.4 & 84.5 & 87.1 & 66.1 \\
InternVL3-8B & 73.6 & 84.0 & 69.5 & 55.4 & 74.2 & 54.4 & 87.5 & 90.5 & 73.3 \\
VL-Rethinker-7B & 68.3 & 75.2 & 60.5 & 50.2 & 75.5 & 55.1 & 77.4 & 86.8 & 65.9 \\
Metis-RISE-7B & 68.4 & 83.7 & 65.9 & 59.3 & 75.8 & 54.9 & 84.2 & 64.5 & 58.5 \\
\midrule
Baseline & 70.3 & 85.0 & 64.7 & 54.8 & 66.8 & 51.5 & 85.0 & 88.4 & 66.1 \\
Baseline+RL & 67.2 & 83.2 & 65.7 & 53.6 & 75.8 & 55.9 & 84.3 & 61.5 & 57.6 \\
\model & 71.2 & 82.5 & 65.4 & 55.7 & 76.0 & 50.4 & 85.0 & 87.0 & 67.6 \\
\bottomrule
\end{tabular}}
\end{table}

\begin{table}[htbp]
\centering
\caption{Performance comparison of our \model against the Baseline$^\dagger$, which is initiated with the original dense instruct model (i.e., Qwen2.5-VL-7B).}
\label{tab:results_diff_init}

\resizebox{\textwidth}{!}{
\setlength{\tabcolsep}{3pt}
\begin{tabular}{lccccccccc}
\toprule
\textbf{Model} & \textbf{Avg.} & \textbf{MMBench} & \textbf{MMStar} & \textbf{MMMU} & \textbf{MathVista} & \textbf{Hallusion} & \textbf{AI2D} & \textbf{OCRBench} & \textbf{MMVet} \\
\midrule
Baseline$^\dagger$ & 70.8 & 81.9 & 66.5 & 56.9 & 74.8 & 49.9 & 83.3 & 85.9 & 67.1  \\
\model & 71.2 & 82.5 & 65.4 & 55.7 & 76.0 & 50.4 & 85.0 & 87.0 & 67.6 \\
\bottomrule
\end{tabular}}

\vspace{1em} %

\resizebox{0.9\textwidth}{!}{
\setlength{\tabcolsep}{3pt}
\begin{tabular}{lccccccc}
\toprule
\textbf{Model} & \textbf{Avg.} & \textbf{MathVista} & \textbf{MathVision} & \textbf{MathVerse} & \textbf{DynaMath} & \textbf{WeMath} & \textbf{LogicVista} \\
\midrule
Baseline$^\dagger$ & 45.4 & 74.8 & 29.5 & 46.7 & 24.8 & 46.4 & 50.1 \\
\model &46.1 & 76.0 & 29.5 & 47.7 & 26.4 & 45.6 & 51.5 \\
\bottomrule
\end{tabular}}

\end{table}

\end{document}